\documentclass[10pt,letterpaper,logo,twocolumn]{style}

\usepackage[numbers]{natbib}
\usepackage{graphicx}
\usepackage{booktabs}
\usepackage{amsmath,amsfonts,amssymb}
\usepackage{subcaption}
\usepackage{multirow}
\usepackage{colortbl}
\usepackage{listings}
\usepackage{xparse}
\usepackage{fontawesome5}
\usepackage{threeparttable}

\graphicspath{{./}{fig/}{figures/}{plot/}{pdf/}{table/}}
\usepackage{amsthm}
\usepackage{tcolorbox}
\tcbuselibrary{skins,breakable}
\tcbuselibrary{listingsutf8}
\usetikzlibrary{arrows.meta}
\usepackage{titletoc}
\usepackage{pifont}
\usepackage{mathtools}
\usepackage{bbm}
\usepackage{makecell}
\usepackage{adjustbox}
\usepackage{xurl}

\title{On-Policy Transition Distillation}
\runningtitle{On Policy Transition Distillation}
\PublicDate{2026-10-19}

\author{%
  {\Authfont
    \textbf{Runyang You}\textsuperscript{1} \quad
    \textbf{Yongqi Li}\textsuperscript{1}\advisor \quad 
    \textbf{Tiezheng Yu}\textsuperscript{2} \quad
    \textbf{Wenjie Li}\textsuperscript{1}
  }\\
  {\Affilfont
    \textsuperscript{1} The Hong Kong Polytechnic University \quad
    \textsuperscript{2} Huawei Technologies Ltd. \\
    \texttt{runyang.y@outlook.com, liyongqi0@gmail.com}
  }
}

\begin{document}

\begin{abstract}

On-policy distillation (OPD) consolidates expert capabilities into a single student by matching teacher and student final-layer distributions on student-generated rollouts.
We investigate whether depthwise prediction transitions, the changes in predictions between layers, can provide additional task-relevant supervision.
Our diagnostic study finds that stronger teacher--student agreement in intermediate transitions is associated with higher answer correctness within questions, with correct responses exhibiting lower mean transition KL across all measured intermediate layer pairs.
These observations motivate On-Policy Transition Distillation (OPTD), which augments final-layer distribution matching by aligning teacher and student layer-to-layer log-probability increments on student-generated rollouts.
Evaluations across three backbones and eight benchmarks spanning mathematics, coding, and instruction following show that OPTD achieves the best or tied-best overall performance in both dual- and three-domain distillation.
In the three-domain setting, OPTD recovers $\approx 90\%$ of the overall teacher--student capability gap, compared with $\approx 75\%$ for OPD.


\end{abstract}


\newcommand{\TitleLinks}{%
\centering
    \vspace{6pt}
    {\noindent\absfont\fontsize{10}{11}\selectfont
    \faGithub\ Project Page: \url{https://github.com/ModalityDance/optd}\par}%
}

\begin{teaserfigure}
  \includegraphics[width=\textwidth]{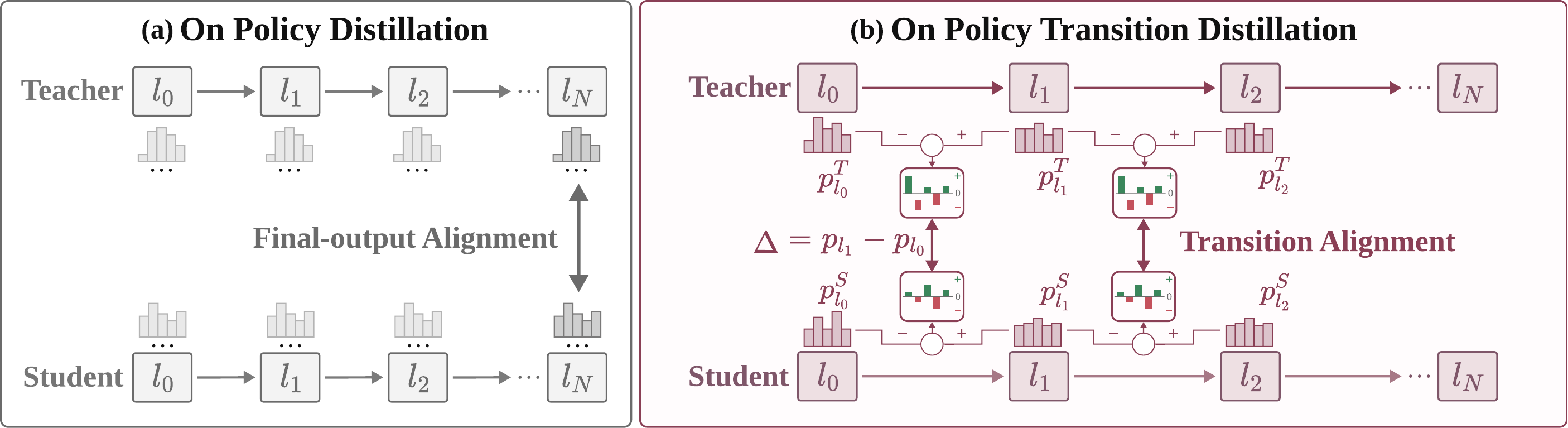}
  \caption{\textbf{On-policy transition distillation.} (a)~Standard OPD aligns only the final-layer predictions of the student and teacher.
    (b)~OPTD additionally aligns their prediction transitions across selected layer segments, transferring how predictions are refined with depth.
}
\label{fig:optd-main}
\end{teaserfigure}

\maketitle

\section{Introduction}
\label{sec:intro}

On-policy distillation (OPD) has emerged as a core post-training recipe for consolidating specialized capabilities from expert teachers into a unified student, as exemplified by DeepSeek-V4 and GLM-5~\citep{deepseek2026v4,glm5team2026glm5}.
Specifically, it applies dense supervision at every token position of a student-generated rollout by comparing the teacher's and student's distributions conditioned on the same prefix, and uses the resulting KL-derived token-level reward to guide policy distillation~\citep{agarwal2024gkd}.
Grounding supervision in the student's own rollouts, OPD alleviates exposure bias~\citep{li2026rethinking}, mitigates catastrophic forgetting~\citep{glm5team2026glm5}, and learns with substantially fewer samples than outcome-reward RL~\citep{yang2025qwen3}.


OPD methods typically transfer knowledge by solely aligning the student's final-layer prediction distribution with those of the teacher.
The final-layer distribution, however, represents only the endpoint of the teacher's decision process. 
Recent studies show that these predictions develop  through successive prediction revisions
across depth: early layers can favor statistical guesses that deeper
layers refine as contextual information develops
~\citep{belrose2023tunedlens,gupta2025llmdepth}.
Layer-contrastive decoding further demonstrates that differences
between predictions across depths, which we term \emph{depthwise prediction transitions}, can be exploited to improve generation factuality~\citep{chuang2024dola}.
This raises a natural question: \emph{can such transitions provide additional task-relevant guidance for on-policy distillation}?

To answer this question, we start by comparing teacher--student agreement on correct and incorrect student-generated responses, and found that \textbf{stronger teacher--student agreement in intermediate prediction transitions is associated with higher answer correctness}, while final-layer agreement exhibits the opposite trend (Fig.~\ref{fig:diagnostic-agreement}(a)).
This correctness-related pattern extends across all measured intermediate layer pairs: correct responses exhibit lower mean transition KL, with larger relative gaps for transitions involving middle-to-late layers (Fig.~\ref{fig:diagnostic-agreement}(b)).
Together, these findings motivate distilling the teacher's internal predictive changes alongside its final output distribution.

We therefore propose
\textbf{On-Policy Transition Distillation (OPTD)}, which complements
final-layer supervision with alignment of teacher--student predictive
transitions on student-generated rollouts, as illustrated in Fig.~\ref{fig:optd-main}(b).
Rather than aligning only the final-layer distributions as in Fig.~\ref{fig:optd-main}(a), OPTD examines
the prediction changes between selected anchor layers for the same
student-visited prefix.
For each candidate token, these changes are represented by the
layer-to-layer increment in its log probability, capturing both the
direction and magnitude of the prediction revision across depth.
We then compare the teacher's and student's increments and use their
discrepancy as a reward alongside the final-layer OPD objective,
encouraging agreement with the teacher's predictive changes on the
same student-generated trajectory.

Evaluations across three model backbones, three domains, and eight benchmarks show that OPTD achieves the best or tied-best overall performance in both dual- and three-domain distillation.
Under matched training budgets, it consistently improves overall performance over standard OPD across all evaluated backbones and domain settings, supporting the joint transfer of heterogeneous specialist capabilities into a single student.
In the three-domain setting, OPTD recovers $\approx 90\%$ of the overall teacher--student capability gap, compared with $\approx 75\%$ for OPD (Tab.~\ref{tab:multidomain-main}).
Internal diagnostics further show that transition supervision improves alignment with the teacher across depth, including intermediate-layer predictions and the final output.

\noindent Our contributions are threefold:
\begin{itemize}[leftmargin=*,noitemsep,topsep=2pt]
    \item We characterize the relationship between teacher–student agreement and answer correctness across model depth, identifying a correctness-related pattern in intermediate predictive transitions that motivates transition-based supervision.
    \item We propose On-Policy Transition Distillation (OPTD), which transfers the teacher's depthwise prediction refinement by aligning the direction and magnitude of cross-layer prediction increments on student-generated rollouts.
    \item We demonstrate that OPTD achieves optimal overall performance in both dual- and three-domain distillation, consistently improving over OPD across backbones and domain settings.
\end{itemize}

\section{Preliminaries and Motivating Observations}
\label{sec:preliminaries}

We first review on-policy distillation (OPD) in \S\ref{subsec:opd}.
We then revisit the assumption underlying standard OPD supervision and compare final-layer output agreement with depthwise transition agreement in \S\ref{subsec:diagnostic-metrics}.
Empirical findings are summarized in \S\ref{subsec:preliminary-analysis}.


\subsection{On-Policy Distillation}
\label{subsec:opd}
OPD trains a student $\pi^S_\theta$ using teacher feedback on its own trajectories.
For a prompt $x\sim\mathcal{D}$, the student generates a response
$y=(y_1,\ldots,y_T)\sim\pi^S_\theta(\cdot\mid x)$ of length $T$.
In multi-domain OPD, prompts from different task domains are supervised by their respective specialist teachers, with $\pi^T$ denoting the teacher for the current prompt~\citep{ma2026mopd}.
At each prefix $s_t=(x,y_{<t})$, where $y_{<t}=(y_1,\ldots,y_{t-1})$ denotes the previously generated tokens.
Let $v\in\mathcal{V}$ denote a candidate next token.
Over the full vocabulary $\mathcal{V}$, the fixed supervision policy $\pi^T$ supplies the token-level reward
\[
R_t(v)
 =\log \pi^T(v\mid s_t)
  -\log \pi^S_{\theta}(v\mid s_t).
\]
At a fixed prefix, the expected reward equals the negative student-to-teacher reverse KL:
\begin{equation}
\label{eq:reverse-kl}
\resizebox{\linewidth}{!}{$\displaystyle
\begin{aligned}
\mathbb{E}_{v\sim\pi^S_\theta(\cdot\mid s_t)}[R_t(v)]
&=\sum_v \pi^S_\theta(v\mid s_t)
\log\frac{\pi^T(v\mid s_t)}{\pi^S_\theta(v\mid s_t)}\\
&=-D_{\mathrm{KL}}\!\left(
\pi^S_\theta(\cdot\mid s_t)
\,\middle\|\,
\pi^T(\cdot\mid s_t)
\right).
\end{aligned}
$}
\end{equation}
Maximizing this reward therefore encourages the student's predictions to match the teacher's on the student's own response prefixes.
OPD implements this objective through the reward-weighted update
\begin{equation}
\label{eq:opd-gradient}
\resizebox{\linewidth}{!}{$\displaystyle
\nabla_{\theta}J_{\mathrm{OPD}}
 =\mathbb{E}_{x\sim\mathcal{D},\,y\sim\pi^S_{\theta}}
  \left[\sum_{t=1}^{T}R_t(y_t)\,
  \nabla_{\theta}\log\pi^S_{\theta}(y_t\mid s_t)\right].
$}
\end{equation}
Following \citet{fu2026revisiting}, we evaluate this update on the student's top-$K$ candidates
\(
\mathcal{K}_t=\operatorname{TopK}\!\left(\pi^S_{\theta}(\cdot\mid s_t)\right).
\)
Renormalizing their probabilities defines the truncated weights
\begin{equation}
\label{eq:opd-topk-weight}
w_t(v)
 =\frac{\pi^S_{\theta}(v\mid s_t)}
 {\sum_{u\in\mathcal{K}_t}\pi^S_{\theta}(u\mid s_t)},
 \qquad v\in\mathcal{K}_t,
\end{equation}
and the resulting OPD loss is
\begin{equation}
\label{eq:opd-topk-loss}
\resizebox{\linewidth}{!}{$\displaystyle
\ell_{\mathrm{OPD},t}
 =-\sum_{v\in\mathcal{K}_t}
 w_t(v)\,R_t(v)\,\log\pi^S_{\theta}(v\mid s_t).
 $}
\end{equation}

\subsection{Revisiting Teacher Agreement in On-Policy Distillation}
\label{subsec:preliminary-study}

\begin{figure}[t]
  \centering
  \includegraphics[width=\linewidth]{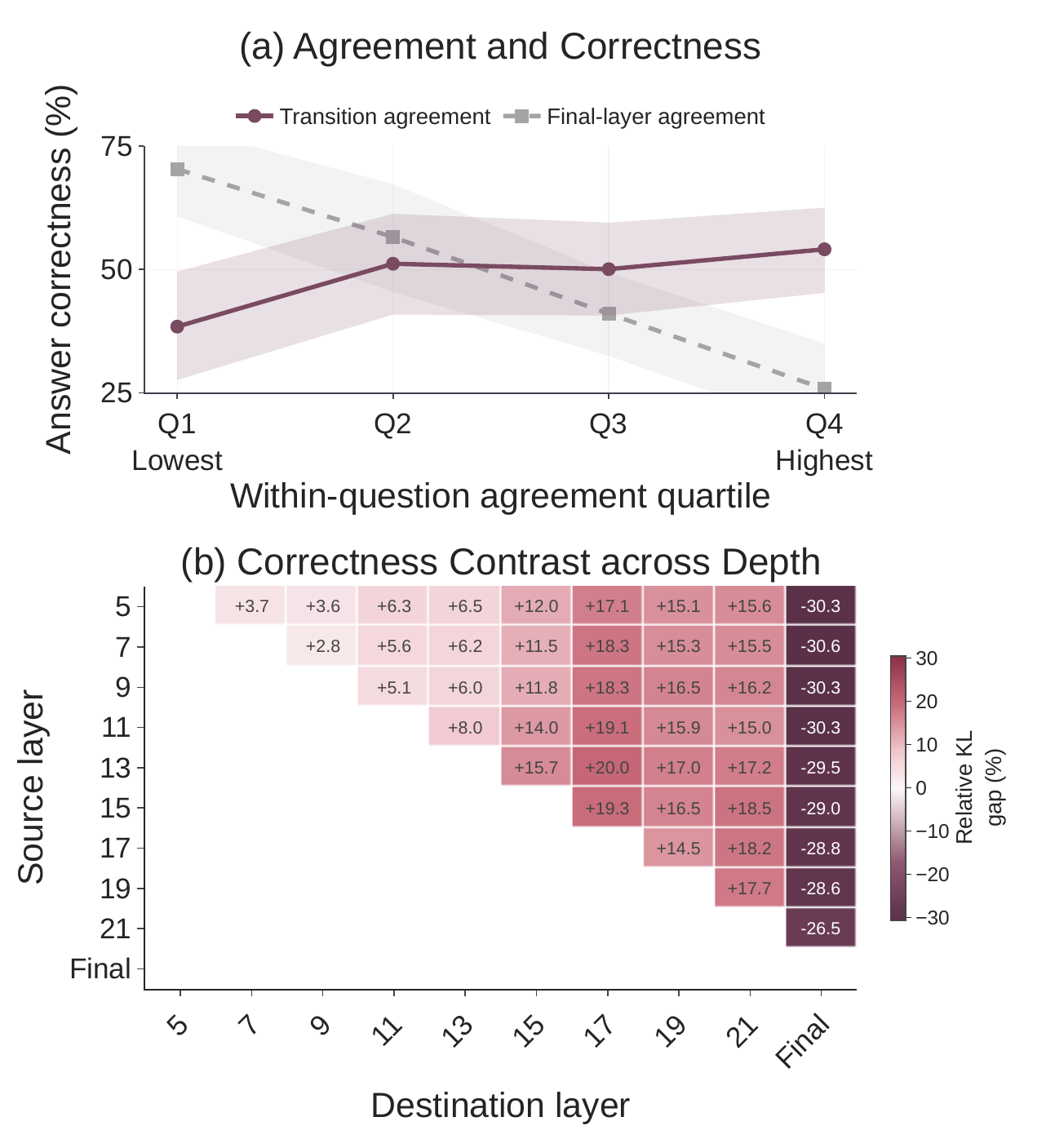}
  \caption{\textbf{Teacher agreement and answer correctness.}
  (a)~Correctness by within-question agreement quartile; shading shows 95\% confidence intervals.
  (b)~Relative transition-KL gaps across layer pairs; positive values indicate closer teacher agreement on correct responses.
  Diagnostic details are in App.~\ref{app:transition-diagnostic}.}
  \label{fig:diagnostic-agreement}
  \label{fig:final-kl-correctness}
  \label{fig:transition-kl-correctness}
\end{figure}

Predictions are progressively refined across transformer layers~\citep{belrose2023tunedlens,gupta2025llmdepth}.
We investigate whether teacher--student agreement in these predictive changes is associated with answer correctness, and how the relationship varies across depth.
We use final-layer agreement as a reference.

\subsubsection{Diagnostic Setup}
\label{subsec:diagnostic-metrics}

\paragraph{Final-Layer Output Divergence.}
We quantify final-layer teacher agreement using the trajectory-averaged reverse KL divergence:
\begin{equation}
\label{eq:metric-final-kl}
\mathcal{D}_{\mathrm{final}}(y)
 =\frac{1}{T}\sum_{t=1}^{T}
 D_{\mathrm{KL}}\!\left(
 \pi^S_{\theta}(\cdot\mid s_t)
 \,\middle\|\,
 \pi^T(\cdot\mid s_t)
 \right).
\end{equation}

\paragraph{Depthwise Transition Divergence.}
A model's final prediction represents the culmination of layered computation.
Across network layers, token predictions undergo progressive revisions, forming \textit{depthwise prediction transitions}.
Let $p^S_{\ell,t}$ and $p^T_{\ell,t}$ denote the intermediate next-token probability distributions obtained by applying the unembedding head to the layer-$\ell$ hidden states (i.e., the logit lens) for the student and teacher, respectively, where $\ell \in \{1,\ldots,L\}$ and $L$ denotes the final layer.
For two layers $\ell < \ell'$, we characterize the revision in candidate log-probabilities by the prediction increment:
\begin{equation}
\label{eq:prediction-increment}
\Delta^{S/T}_{\ell\rightarrow\ell',t}(v)
 =\log p^{S/T}_{\ell',t}(v) - \log p^{S/T}_{\ell,t}(v).
\end{equation}
To evaluate whether the student follows the teacher's internal revision trajectory over the vocabulary $\mathcal{V}$, we normalize the increments into a probability distribution:
\begin{equation}
\label{eq:normalized-increment}
q^{S/T}_{\ell\rightarrow\ell',t}(v)
 =\frac{\exp(\Delta^{S/T}_{\ell\rightarrow\ell',t}(v))}
 {\sum_{j\in\mathcal{V}}\exp(\Delta^{S/T}_{\ell\rightarrow\ell',t}(j))}.
\end{equation}
This distribution assigns higher probability mass to candidate tokens whose probabilities increase by a larger factor between the layers.
We then measure transition divergence between layers $\ell$ and $\ell'$ using response-averaged reverse KL:
\begin{equation}
\label{eq:metric-trans-kl}
\mathcal{D}_{\mathrm{trans}}(\ell,\ell';y)
 =\frac{1}{T}\sum_{t=1}^{T}
 D_{\mathrm{KL}}\!\left(
 q^{S}_{\ell\rightarrow\ell',t}
 \,\middle\|\,
 q^{T}_{\ell\rightarrow\ell',t}
 \right).
\end{equation}
A lower value of $\mathcal{D}_{\mathrm{trans}}(\ell,\ell';y)$ indicates that the student's internal prediction transitions follow the teacher's revision trajectory more faithfully.

\paragraph{Comparison Protocol.}
We analyze 480 responses from 60 mathematics questions that elicit both correct and incorrect answers from an OPD-trained Qwen3-4B student.
To compare responses to the same question, we rank its answers by transition agreement, defined as the negative mean divergence over the measured intermediate layer pairs, and report correctness across four agreement groups (Fig.~\ref{fig:diagnostic-agreement}(a)).
We repeat this ranking using negative final-layer KL as a reference.
To examine the depthwise pattern, we compare correct and incorrect responses at each layer pair (Fig.~\ref{fig:diagnostic-agreement}(b)).
Let $\overline{\mathcal{D}}_{\mathrm{correct}}$ and $\overline{\mathcal{D}}_{\mathrm{incorrect}}$ denote the respective group means, aggregated within questions and then across benchmarks.
We report their relative gap:
\begin{equation}
\label{eq:relative-divergence-gap}
\Delta \mathcal{D} = \frac{\overline{\mathcal{D}}_{\mathrm{incorrect}} - \overline{\mathcal{D}}_{\mathrm{correct}}}{\overline{\mathcal{D}}_{\mathrm{correct}}} \times 100\%.
\end{equation}
Positive values indicate closer teacher agreement on correct responses.
The checkpoint, layer pairs, aggregation, and uncertainty estimation are detailed in App.~\ref{app:transition-diagnostic}.

\subsubsection{Observation and Analysis}
\label{subsec:preliminary-analysis}

\paragraph{Observation 1: Agreement in predictive transitions tracks answer correctness.}
Agreement with the teacher's intermediate predictive changes carries a task-relevant signal: among responses to the same question, those with stronger transition agreement tend to be more accurate (Fig.~\ref{fig:diagnostic-agreement}(a)).
Correctness rises from 38.4\% in the lowest-agreement quartile to 54.1\% in the highest, with similar rates in the two middle quartiles.
This association links response correctness to how closely the student follows the teacher's predictive changes across depth, motivating transitions as a candidate distillation target.
Final-layer agreement exhibits the opposite trend in the same response cohort.

\paragraph{Observation 2: The correctness contrast is structured across depth.}
\label{subsec:transition-diagnostic}
The correctness-related contrast extends across all 36 measured intermediate layer pairs: correct responses exhibit lower mean transition KL than incorrect responses (Fig.~\ref{fig:diagnostic-agreement}(b)).
Within this internal grid, the relative gaps are more pronounced for transitions involving middle-to-late layers.
Transitions ending at the final output exhibit the opposite direction, distinguishing the internal pattern from the changes leading to the output distribution.
This depthwise structure is consistent with accounts of predictions being progressively refined as contextual information develops across transformer layers~\citep{belrose2023tunedlens,gupta2025llmdepth}.
It motivates examining the teacher's internal predictive changes as a task-relevant aspect of its behavior.

Together, these findings motivate distilling the teacher's predictive changes across depth in addition to its final output distribution.
We therefore introduce OPTD, which augments final-layer distribution matching with supervision on depthwise prediction transitions.

\section{On-Policy Transition Distillation}
\label{sec:method}

Building on the observations, we propose \textbf{On-Policy Transition Distillation (OPTD)} to distill the teacher's depthwise prediction increments from on policy rollouts.

\label{subsec:transition-objective}
Consider an ordered sequence of anchor layers $\ell_0 < \ell_1 < \dots < \ell_M=L$.
For each segment $i$, from $\ell_{i-1}$ to $\ell_i$, the prediction increment $\Delta^{S/T}_{i,t}(v)$ is the quantity defined in Eq.~\eqref{eq:prediction-increment}.
Following the top-$K$ setting introduced in Section~\ref{sec:preliminaries}, we apply the same reward-weighted update to these depthwise increments.
The transition-level reward is
\begin{equation}
\label{eq:transition-reward}
R^{\mathrm{trans}}_{i,t}(v)
=
\Delta^T_{i,t}(v)-\Delta^S_{i,t}(v).
\end{equation}
For the segment ending at $\ell_i$, the student probabilities of the retained candidates are renormalized at the destination layer:
\begin{equation}
\label{eq:transition-weight}
w_{i,t}(v)
 =\frac{p^S_{\ell_i,t}(v)}
 {\sum_{u\in\mathcal{K}_t}p^S_{\ell_i,t}(u)},
 \qquad v\in\mathcal{K}_t.
\end{equation}
Using the detached transition signal and weights, the transition loss is
\begin{equation}
\label{eq:transition-gradient}
\resizebox{\linewidth}{!}{$\displaystyle
\mathcal{L}_{\mathrm{trans}}
 =-\mathbb{E}_{x\sim\mathcal{D},\,y\sim\pi^S_\theta}
 \!\left[
 \mathbb{E}_{t}\left[
 \frac{1}{M} \sum_{i=1}^{M}
 \sum_{v\in\mathcal{K}_t}
 w_{i,t}(v)\,R^{\mathrm{trans}}_{i,t}(v)\,
 \Delta^S_{i,t}(v)
 \right]
 \right].
$}
\end{equation}
During training, we retain the inherited final-layer update and add the transition update above $\nabla_\theta J = \nabla_\theta J_{\mathrm{OPD}} + \nabla_\theta J_{\mathrm{trans}}$.
The relation between this practical update and the exact reverse transition-KL gradient is derived in App.~\ref{app:reverse-transition-kl}; implementation details are given in App.~\ref{app:opd-transition-implementation}.

\section{Experiment Setup}
\label{sec:experiment-setup}
\label{subsec:exp-benchmarks-models}

We evaluate OPTD in dual-domain (\S\ref{subsec:dual-domain-results}) and three-domain (\S\ref{subsec:multi-teacher-generalization}) distillation settings, covering mathematics and coding, with instruction following added in the latter.
Each prompt receives supervision from its corresponding domain specialist, following the protocol described in\S\ref{subsec:opd}.

\paragraph{Datasets and Benchmarks.}
We train on DAPO-Math-17k~\citep{yu2025dapo} and DeepCoder-Preview-Dataset~\citep{luo2025deepcoder}.
We evaluate mathematics on AIME24/25~\citep{maa2025aime} and AIMO~\citep{aimo2024validation}; coding on LiveCodeBench v5~\citep{jain2024livecodebench}, CodeContests~\citep{li2022alphacode}, and Codeforces with CodeElo~\citep{quan2025codeelo}.
For instruction following, we use Rubrics-To-Tokens~\citep{turleingrubrics} for training and IFEval~\citep{zhou2023ifeval} for evaluation.
Training prompts are sampled equally across domains.

\paragraph{Models.}
We use DeepSeek-R1-Distill-Qwen-1.5B/7B~\citep{guo2025deepseekr1} and Qwen3-4B~\citep{yang2025qwen3} as students, each supervised by domain specialists detailed in App.~\ref{app:experiment-models}.

\paragraph{Baselines.}
We compare OPTD with OPD and its latent-supervision variants, OPRD~\citep{yang2026oprd} and LastOPD~\citep{yang2026lastopd}, which transfer teacher knowledge through hidden representations.
We also implement LayerOPD: an \emph{absolute-layer supervision} baseline that applies OPD across corresponding intermediate layer readouts.

\paragraph{Training.}
Methods within each setting share the student initialization, teachers, data, and update budget: 150 updates for 1.5B, 50 for 4B, and 100 for 7B.
We use a learning rate of $10^{-5}$ and two on-policy responses per prompt.
Qwen3-4B uses non-thinking mode, while the R1-Distill students use reasoning prompts.

\paragraph{Evaluation.}
We report Avg@16 for mathematics, Avg@8/Pass@8 for LiveCodeBench, Avg@16 for CodeContests, estimated Codeforces Elo, and average strict/loose IFEval accuracy under 16 samples per question.
Dataset links, baseline implementations, and detailed training and evaluation settings are provided in App.~\ref{app:experiment-details}.

\definecolor{TableGroup}{HTML}{F1F3F5}
\definecolor{TableAccent}{HTML}{951f2f}
\definecolor{TableOurs}{HTML}{fff0f2}
\definecolor{TableAccentGrey}{HTML}{755459}
\definecolor{DarkRose}{HTML}{7a4a62}
\begin{table*}[t]
    \centering
    \resizebox{\textwidth}{!}{
    \begin{tabular}{l*{7}{c}*{8}{c}c}
    \toprule
    & \multicolumn{7}{c}{Math} & \multicolumn{8}{c}{Coding} & \\
    \cmidrule(lr){2-8}\cmidrule(lr){9-16}
    Method & \multicolumn{2}{c}{AIME24} & \multicolumn{2}{c}{AIME25}
      & \multicolumn{2}{c}{AIMO} & \multirow{2}{*}{Avg.}
      & \multicolumn{2}{c}{LiveCodeBench} & \multicolumn{2}{c}{CodeContests}
      & \multicolumn{3}{c}{CodeElo} & \multirow{2}{*}{Avg.} & \textbf{Overall} \\
    \cmidrule(lr){2-3}\cmidrule(lr){4-5}\cmidrule(lr){6-7}
    \cmidrule(lr){9-10}\cmidrule(lr){11-12}\cmidrule(lr){13-15}
    & Avg & Pass & Avg & Pass & Avg & Pass &
      & Avg & Pass & Avg & Pass & Avg & Pass & Elo & & \\
   \midrule
   \rowcolor{TableGroup}\multicolumn{17}{l}{\texttt{DeepSeek-R1-Distill-Qwen-1.5B}}\\ 
   OPD   & \textbf{47.29}& 80.00 & 30.00 & 53.33& 74.40 & 91.57& \underline{50.56} & \textbf{22.98} & 37.63 & 9.58   & \underline{25.45}   & 8.55    & 19.36     & 843.26 & 13.70 & 32.13 \\
   LastOPD & 43.75 & \underline{76.67} & \underline{31.46} & 56.67 & \underline{75.90} & \underline{93.98} & 50.37 & {22.40} & \underline{38.35} & \textbf{10.57} & 26.06 & \underline{9.87} & \underline{21.57} & 916.11 & \textbf{14.28} & \underline{32.33} \\
   OPRD  & 36.88        & 73.33        & 27.50      & 53.33 & 71.46 & 91.57& 45.28   & 20.03 & 34.77 & 9.58   & 22.42   & 8.55  & 19.36  & 843.26 & 12.72 & 29.00        \\
   LayerOPD   & 41.67        & 76.67      & 30.00      & 50.00  & \textbf{76.73}      & 93.98  & 49.47   & 22.40 & 36.92 & 9.62 & 23.64 & \textbf{10.26} & \textbf{22.55} & \underline{931.23} & \underline{14.09} & 31.78        \\
   \rowcolor{TableOurs}
   \textcolor{TableAccent}{\textbf{OPTD} (Mid)} & \underline{45.42}        & 73.33        & \textbf{31.88}  & \textbf{56.67}       & 75.75 & \textbf{95.18}     & \textbf{51.02}   & 22.04 & 36.20 & 9.77   & 23.64   & 9.28  & 19.85  & 852.37 & 13.70 & \textbf{32.36}        \\
   \rowcolor{TableOurs}
   \textcolor{TableAccent}{\textbf{OPTD} (Late)}   & 44.17        & \textbf{80.00}& 31.04 & \underline{53.33}    & 74.55 & 91.57& 49.92   & \underline{22.40} & \textbf{39.07} & \underline{10.11}  & \textbf{26.06}   & 9.56  & \underline{21.81}  & \textbf{953.25} & 14.02 & 31.97        \\
   \midrule
   \rowcolor{TableGroup}\multicolumn{17}{l}{\texttt{Qwen3-4B (non-thinking)}} \\ 
   OPD   & 43.75        & 73.33 & 31.04      & 66.67    & 73.42 & 91.57  & 49.40   & \underline{32.97} & 48.75 & 7.99   & 27.27   & 13.42 & \textbf{28.43}  & \textbf{1168.12} & 18.13 & 33.77        \\
   LastOPD & 41.88 & 70.00 & 30.21 & 60.00 & 70.86 & 89.16 & 47.65 & \textbf{33.65} & \textbf{49.82} & \underline{12.88} & \underline{32.12} & \textbf{13.79} & \underline{27.45} & 1156.18 & \textbf{20.11} & 33.88 \\
   OPRD    & 45.00  & 73.33 & 31.88 & 66.67 & 73.57 & 90.36 & 50.15 & 29.66 & 44.32 & \textbf{12.99} & \textbf{35.15} & 13.14 & 26.96 & \underline{1158.86} & \underline{18.60} & \underline{34.38} \\
   LayerOPD   & \underline{46.25}  & \textbf{76.67} & \underline{32.50} & 60.00 & \underline{73.95} & \underline{92.77} & \underline{50.90} & 30.02 & 43.71 & 12.50 & \textbf{35.15} & 12.87 & 25.98 & 1107.25 & 18.46 & 34.68 \\
   \rowcolor{TableOurs}
   \textcolor{TableAccent}{\textbf{OPTD} (Mid)} & 45.63 & \underline{73.33} & 32.29      & \underline{66.67}    & 73.42 & 90.36& 50.45   & 31.50 & \underline{48.75} & 8.67   & 31.52   & \underline{13.69} & 25.98  & 1110.04 & 17.95 & 34.20        \\
   \rowcolor{TableOurs}
   \textcolor{TableAccent}{\textbf{OPTD} (Late)}   & \textbf{47.29}& 70.00        & \textbf{33.96}        & \textbf{70.00}       & \textbf{74.77}   & \textbf{93.98}     & \textbf{52.01}   & 30.73 & 44.80 & 8.48     & 28.48      & 12.84    & 26.47     & 1075.77 & 17.35 & \textbf{34.68}        \\
   \midrule
   \rowcolor{TableGroup}\multicolumn{17}{l}{\texttt{DeepSeek-R1-Distill-Qwen-7B}} \\
   OPD   & 62.92 & \underline{80.00} & 47.71 & \underline{76.67} & 86.75 & \underline{96.39} & 65.79 & 47.76 & \underline{62.01} & \underline{23.60} & \underline{45.45} & 26.07 & 47.30 & \underline{1776.93} & 32.48 & 49.14 \\
   LastOPD& 63.54 & 83.33 & \underline{48.96} & \textbf{80.00} & \underline{87.50} & 95.18 & 66.67 & \underline{48.30} & 60.22 & 22.80 & 43.03 & \textbf{27.48} & \textbf{49.51} & \textbf{1797.07} & \underline{32.86} & 49.77        \\
   LayerOPD   & \textbf{67.50}  & 83.33  & 46.46 & 70.00 & 86.37 & 93.98 & \underline{66.78} & 47.76 & 58.06 & 23.41 & \textbf{47.88} & \underline{27.21} & \underline{49.02} & 1759.54 & 32.79 & \underline{49.79} \\
   \rowcolor{TableOurs}
   \textcolor{TableAccent}{\textbf{OPTD (Late)}}   & \underline{63.54} & \textbf{83.33} & \textbf{49.38} & 70.00 & \textbf{88.10} & \textbf{98.80} & \textbf{67.01} & \textbf{48.39} & \textbf{64.16} & \textbf{23.98} & 44.85 & 26.93 & 48.77 & 1740.09 & \textbf{33.10} & \textbf{50.06} \\
   \bottomrule
        \end{tabular}%
    }
    \caption{Dual-domain distillation results on mathematics and coding benchmarks across model backbones.
        All scores are percentages except CodeElo reports the estimated rating.
        Highest and second-highest values in each column are bold and underlined, respectively.}
    \label{tab:main-results-4b}
\end{table*}

\section{Main Results}
\label{sec:main-results}

\subsection{Results on Dual-Domain Setting}
\label{subsec:dual-domain-results}

Table~\ref{tab:main-results-4b} evaluates whether depthwise transition
supervision improves the joint transfer of mathematics and coding
capabilities under matched training settings.
\begin{itemize}[leftmargin=*,noitemsep,topsep=2pt]
\item \textbf{Transition supervision improves overall capability transfer across model sizes.}
The best OPTD configuration achieves the highest Overall score at each of
the three model sizes, tying LayerOPD at 4B.
At 4B, for example, Late OPTD raises Overall from 33.77\% to 34.68\%
and improves all three mathematics benchmark averages over OPD.
At 7B, it improves both Math Avg and Coding Avg over OPD, raising Overall
from 49.14\% to 50.06\%.
These results support transition supervision as a useful addition to
final-output distillation for consolidating specialist capabilities in
a single student.
The preferred supervision depth depends on the setting: Middle OPTD leads
at 1.5B, whereas Late OPTD leads at 4B and 7B.

\item \textbf{Representation alignment is effective, but its gains can involve cross-domain tradeoffs.}
Representation-based supervision can enhance capability transfer:
LastOPD improves Overall over OPD at all three sizes and achieves the
highest Coding Avg at 1.5B and 4B, while OPRD improves both domain averages
over OPD at 4B.
However, LastOPD's coding gains at 1.5B and 4B accompany lower Math Avg
than OPD in both settings.
This pattern illustrates a cross-domain tradeoff: stronger transfer in
one domain does not necessarily translate into stronger transfer in the other.
For multi-specialist distillation, these results motivate assessing
representation alignment through joint capability transfer, since a
leading domain-specific score can coexist with a tradeoff elsewhere.
\end{itemize}

\subsection{Results on Three-Domain Setting}
\label{subsec:multi-teacher-generalization}

\newcommand{\rc}[1]{\textcolor{black!65}{\itshape #1}}
\newcommand{\rch}[1]{\textcolor{TableAccentGrey}{\textbf{\textit{#1}}}}
\begin{table*}[t]
    \centering
    \resizebox{\textwidth}{!}{%
    \begin{tabular}{lccccc|cccccc|ccccc|c}
    \toprule
     & \multicolumn{5}{c|}{Math} & \multicolumn{6}{c|}{Coding} & \multicolumn{5}{c|}{Instruction Following} & \\
    \cmidrule(lr){2-6}\cmidrule(lr){7-12}\cmidrule(lr){13-17}
    Method & \multicolumn{2}{c}{AIME24} & \multicolumn{2}{c}{AIME25} & \multirow{2}{*}{Avg.} & \multicolumn{2}{c}{LiveCodeBench} & \multicolumn{2}{c}{CodeContests} & CodeElo & \multirow{2}{*}{Avg.} & \multicolumn{2}{c}{IFEval} & \multicolumn{2}{c}{IFBench} & \multirow{2}{*}{Avg.} &  \textbf{Overall} \\
    \cmidrule(lr){2-3}\cmidrule(lr){4-5}\cmidrule(lr){7-8}\cmidrule(lr){9-10}\cmidrule(lr){11-11}\cmidrule(lr){13-14}\cmidrule(lr){15-16}
     & Avg & Pass & Avg & Pass & & Avg & Pass & Avg & Pass & Elo & & prompt & instr. & prompt & instr. & & \\
    \midrule
    Student & 30.63 & 70.00 & 21.04 & 33.33 & 25.84 & 14.43 & 27.96 & 5.15 & 21.82 & 618.23 & 9.79 & 41.77 & 53.59 & 13.84 & 15.70 & 31.23 & 29.11 \\
    Teachers & 48.96 & 76.67 & 34.79 & 56.67 & 41.88 & 21.86 &36.56 & 10.11 & 23.03 & 943 & 15.99 & 69.67 & 77.90 & 16.00 & 17.74 & 45.33 & 40.83 \\
    \midrule
    OPD & 43.54 & 73.33 & 30.63 & 50.00 & 37.09 & \textbf{22.45} & 36.56 & 10.57 & \textbf{26.06} & 946.11 & \textbf{16.51} & 61.54 & 69.74 & 13.84 & 16.42 & 40.39 & 37.89 \\
    \rc{recovery rate} & \rc{70.4\%} & \rc{49.9\%} & \rc{69.7\%} & \rc{71.4\%} & \rc{70.1\%} & \rch{107.9\%} & \rc{100.0\%} & \rc{109.3\%} & \rch{350.0\%} & \rc{100.9\%} & \rch{108.4\%} & \rc{70.9\%} & \rc{66.4\%} & \rc{0.0\%} & \rc{35.3\%} & \rc{65.0\%} & \rc{74.9\%} \\
    \rowcolor{TableOurs}\textcolor{TableAccent}{\textbf{OPTD}} & \textbf{46.67} & \textbf{80.00} & \textbf{32.50} & \textbf{53.33} & \textbf{39.59} & 22.13 & \textbf{36.92} & \textbf{10.72} & 25.45 & \textbf{958.12} & 16.43 & \textbf{61.79} & \textbf{72.16} & \textbf{15.67} & \textbf{17.45} & \textbf{41.77} & \textbf{39.57} \\
    \rowcolor{TableOurs} \rch{recovery rate} & \rch{87.5\%} & \rch{149.9\%} & \rch{83.3\%} & \rch{85.7\%} & \rch{85.7\%} & \rc{103.6\%} & \rch{104.2\%} & \rch{112.3\%} & \rc{299.6\%} & \rch{104.6\%} & \rc{107.1\%} & \rch{71.8\%} & \rch{76.4\%} & \rch{84.7\%} & \rch{85.8\%} & \rch{74.8\%} & \rch{89.2\%} \\
    \bottomrule
        \end{tabular}%
    }
    \caption{Three-domain distillation results across mathematics, coding,
        and instruction following on the 1.5B backbone.
        Teachers denotes domain-specialized teachers evaluated on their target domains.}
    \label{tab:multidomain-main}
\end{table*}

We extend distillation to a 1.5B student with specialist teachers for mathematics,
coding, and instruction following.
Table~\ref{tab:multidomain-main} compares OPD and OPTD under identical
update budgets.
\begin{itemize}[leftmargin=*,noitemsep,topsep=2pt]
    \item \textbf{The benefits of transition supervision extend to heterogeneous capabilities.}
    OPTD improves every reported mathematics and instruction-following
    metric over OPD, with instruction-following gains spanning both
    IFEval and IFBench.
    These simultaneous improvements raise Overall from 37.89\% to 39.57\%.
    The results extend the evidence for OPTD beyond math/coding distillation:
    transition supervision also supports the integration of instruction-following
    capabilities alongside specialist reasoning skills in one student.

    \item \textbf{OPTD recovers more of the specialist capabilities left untapped by OPD.}
    OPD already reaches the coding teacher's domain average, while recovering
    70.1\% and 65.0\% of the teacher--student gaps in mathematics and
    instruction following, respectively.
    OPTD increases these recovery rates to 85.7\% and 74.8\%, while retaining
    a coding domain average above the teacher reference.
    Thus, the overall improvement reflects additional transfer in the
    domains with larger remaining teacher gaps, alongside sustained
    specialist-level coding performance.
\end{itemize}
These results support capability transfer from heterogeneous specialists;
complete results appear in App.~\ref{app:multi-teacher-full-results}
and Tabs.~\ref{tab:multidomain-math}--\ref{tab:multidomain-instruction}.

\section{Further Analysis}
\label{sec:analysis}

This section examines the design and mechanism of transition supervision.
We study how transition geometry affect transfer, how alternative
objective formulations compare, and whether OPTD improves the student's
internal prediction alignment with the teacher.
Further diagnostic and derivational analyses are provided in
Apps.~\ref{app:additional-analyses} and \ref{app:reverse-transition-kl}.

\subsection{Transition Geometry Analysis}
\label{subsec:transition-geometry-analysis}


\newcommand{\transitionpathgray}[3]{%
    \begin{tikzpicture}[baseline=-0.55ex,x=1.55cm,y=0.16cm]
        \draw[gray!55,line width=.45pt,-{Stealth[length=4.2pt,width=4pt]}] (0,0)--(1,0);
        \foreach \x in {0,.5}{\draw[gray!55,line width=.3pt] (\x,-.05)--(\x,.05);}
        \draw[DarkRose,line width=.65pt] (#1,0)--(#2,0)--(#3,0);
        \foreach \x in {#1,#2,#3}{\fill[DarkRose] (\x,0) circle (1.6pt);}
    \end{tikzpicture}%
}
\newcommand{\transitionpathblue}[3]{%
    \begin{tikzpicture}[baseline=-0.55ex,x=1.55cm,y=0.16cm]
        \draw[gray!45,line width=.45pt,-{Stealth[length=4.2pt,width=4pt]}] (0,0)--(1,0);
        \foreach \x in {0,.5}{\draw[gray!45,line width=.3pt] (\x,-.05)--(\x,.05);}
        \draw[blue!70!black,line width=1.05pt] (#1,0)--(#2,0)--(#3,0);
        \foreach \x in {#1,#2,#3}{\fill[blue!70!black] (\x,0) circle (1.9pt);}
    \end{tikzpicture}%
}

\newcolumntype{L}[1]{>{\raggedright\arraybackslash}p{#1}}
\newcolumntype{C}[1]{>{\centering\arraybackslash}p{#1}}
\newcolumntype{R}[1]{>{\raggedleft\arraybackslash}p{#1}}

We examine how the placement and span of a supervised transition path
affect capability transfer in the 1.5B multi-domain setting, where Math is the average of AIME24, AIME25, and AIMO;
Coding is LiveCodeBench, and IF is accuracy on IFEval.
We express depth as $r=\ell/L$, where $L=28$ for this backbone,
with the final prediction at $r=1$.
We define \emph{placement} as the normalized depth $c$ of the central
supervised block and \emph{span} as the normalized distance $s$ from
that block to each neighboring block.
Given center $c$ and span $s$, the intermediate readouts are positioned
at $(c-s,c,c+s)$.
Early, Middle, and Late placements have centers at approximately
$0.393$, $0.464$, and $0.536$, respectively.
Under otherwise matched training conditions (\S\ref{sec:experiment-setup}),
we first compare placements at a fixed span of $s\approx0.214$, then compare
spans of approximately $0.143$, $0.214$, and $0.286$ at the Middle and
Late placements.
Exact block indices are provided in App.~\ref{app:experiment-training}.

\begin{table}[t]
    \centering
    \resizebox{\linewidth}{!}{%
        \begin{tabular}{L{1.cm}C{1.9cm}*{4}{C{0.8cm}}}
        \toprule
        Placement   & Path  & Math & Coding & IF & Avg. \\
        \midrule
        Early & \transitionpathgray{.179}{.393}{.607}  & 49.32 & \underline{21.59}      & \underline{64.77} & \underline{45.23} \\
        Middle & \transitionpathgray{.250}{.464}{.679}  & \textbf{49.93} & 20.61      & 64.29 & 44.94 \\
        Late & \transitionpathgray{.321}{.536}{.750} & \underline{49.46} & \textbf{22.67}      & \textbf{64.97} & \textbf{45.70} \\
        \bottomrule
        \end{tabular}%
    }
    \caption{Depth placement comparison at a fixed normalized span of $s\approx0.214$.
        Path glyphs visualize placement along the normalized depth axis $r=\ell/28$.}
    \label{tab:transition-depth}
\end{table}

\begin{table}[t]
    \centering
    \small
    \setlength{\tabcolsep}{3.5pt}
    \resizebox{\linewidth}{!}{%
    \begin{tabular}{L{1.0cm}C{1.9cm}*{4}{C{0.9cm}}}
    \toprule
    Span & Path   & Math  & Coding & IF & Avg. \\
    \midrule
    \rowcolor{TableGroup}\multicolumn{6}{l}{\textit{Middle}} \\
    0.14 & \transitionpathgray{.321}{.464}{.607} & 49.27 & \textbf{22.94}      & \underline{66.16} & \underline{46.12} \\
    0.29 & \transitionpathgray{.179}{.464}{.750} & 48.41 & 21.33      & 65.72 & 45.15 \\
    0.21 & \transitionpathgray{.250}{.464}{.679} & \underline{49.93} & 20.61      & 64.29 & 44.94 \\
    \midrule
    \rowcolor{TableGroup}\multicolumn{6}{l}{\textit{Late}} \\
    0.14 & \transitionpathgray{.393}{.536}{.679} & \textbf{51.16} & 22.13      & \textbf{66.98}  & \textbf{46.76} \\
    0.29 & \transitionpathgray{.250}{.536}{.821} & 48.77 & 21.60      & 65.37 & 45.25 \\
    0.21 & \transitionpathgray{.321}{.536}{.750} & 49.46 & \underline{22.67}      & 64.97 & 45.70 \\
    \bottomrule
    \end{tabular}%
    }
    \caption{Span comparison at Middle and Late placements. Path glyphs visualize placement along the normalized depth axis $r=\ell/28$.}
    \label{tab:transition-span}
\end{table}

\begin{itemize}[leftmargin=*,noitemsep,topsep=2pt]
    \item \textbf{The best placement depends on the target domain.}
    At a fixed span of $s\approx0.214$ (Tab.~\ref{tab:transition-depth}), Middle achieves
    the highest Math score, whereas Late achieves the highest Coding and IF
    scores and the highest overall score.
    Thus, placement affects which capabilities benefit most from transition
    supervision, with Late providing the strongest joint transfer in this comparison.

    \item \textbf{Compact paths achieve the best overall performance.}
    The compact span $s\approx0.143$ achieves the highest average at both Middle and Late
    (Tab.~\ref{tab:transition-span}).
    Relative to $s\approx0.286$, it raises the average by 0.97 at Middle and 1.51 at Late, improving all three domain scores
    at both placements.
    This repeated pattern supports supervising more local prediction
    revisions rather than spanning a broader depth interval in these settings.
\end{itemize}

\noindent The compact Late path achieves the highest overall score in
this sweep (46.76\%), supporting its use in the main three-domain experiment.
Its intermediate readouts lie at approximately $(0.393,0.536,0.679)$,
with center $c\approx0.536$ and span $s\approx0.143$.

\subsection{Objective Formulation Analysis}
\label{subsec:objective-formulation-analysis}


\begin{table}[t]
    \centering
    \resizebox{\columnwidth}{!}{%
        \begin{tabular}{lccrrr}
   \toprule
   Formulation & Weight & Centered & Math & Coding & Avg. \\
   \midrule
   ReverseKL & $q^S$ & \checkmark & 50.78 & 21.15 & 35.97 \\
   \rowcolor{TableOurs}\textcolor{TableAccent}{\textbf{OPTD}} & $w_{i,t}$ & \texttimes & \textbf{51.01} & 21.68 & \textbf{36.35} \\
   \midrule
   \shortstack[l]{Increment weights\\+ raw reward} & $q^S$ & \texttimes & 50.48 & \textbf{21.77} & 36.13 \\
   \shortstack[l]{Predictive weights\\+ centered reward} & $w_{i,t}$ & \checkmark & 43.13 & 0.36 & 21.75 \\
   \bottomrule
        \end{tabular}%
    }
    \caption{Transition update comparison on the 1.5B dual-domain setting.
        Weights use increment softmax $q^S$ or predictive probabilities $w_{i,t}$;
        centering subtracts the $q^S$-weighted mean reward.
        Math is Math Avg@16 and Coding is LiveCodeBench Avg@4.
        Avg.\ is their equal-weight mean; bold marks the best score.}
    \label{tab:transition-objective-ablation}
\end{table}

Our motivating analysis uses transition KL to measure agreement between
normalized prediction increments.
We now examine how to turn this transition signal into a training update,
comparing direct transition-KL minimization with OPTD's OPD-style formulation.
\emph{ReverseKL} uses increment-softmax weights $q^S$ and a centered
transition reward (App.~\ref{app:reverse-transition-kl});
\emph{OPTD} uses destination-layer predictive weights $w_{i,t}$ and the
raw transition reward $R^{\mathrm{trans}}_{i,t}$
(Eqs.~\eqref{eq:transition-reward}--\eqref{eq:transition-gradient}).
Two hybrid variants pair increment weights with the raw reward and
predictive weights with the $q^S$-centered reward, respectively.
We compare all four formulations under controlled conditions in the
1.5B dual-domain setting.

\begin{itemize}[leftmargin=*,noitemsep,topsep=2pt]
    \item \textbf{OPTD improves both domains over direct transition-KL matching.}
    OPTD achieves the highest average in Table~\ref{tab:transition-objective-ablation}
    (36.35\%), improving on ReverseKL by 0.23 percentage points in Math
    and 0.53 points in Coding.
    Its overall gain therefore reflects improvements in both domains,
    supporting the OPD-style transition update for capability transfer
    in this setting.

    \item \textbf{Raw rewards work better with predictive weights.}
    To test OPTD's choice of raw transition rewards, we hold predictive
    weights fixed and introduce reward centering.
    The raw-reward update achieves an average of 36.35\%, whereas the
    centered variant reaches 21.75\%.
    Under increment weights, however, centered and raw rewards yield
    similar averages, locating the
    adverse effect of centering in its pairing with predictive weights
    in this comparison.
    The gradient analysis in App.~\ref{app:centered-predictive-instability}
    further shows very large gradient norms for the centered predictive
    variant, indicating optimization instability during training.
    These results support retaining raw transition rewards when using
    OPTD's predictive weights.
\end{itemize}

\noindent These comparisons support OPTD's pairing of destination-layer
predictive weights with the raw transition reward.

\subsection{Internal Alignment Analysis}
\label{subsec:prediction-path-analysis}

\begin{figure}[t]
    \centering
    \includegraphics[width=\columnwidth]{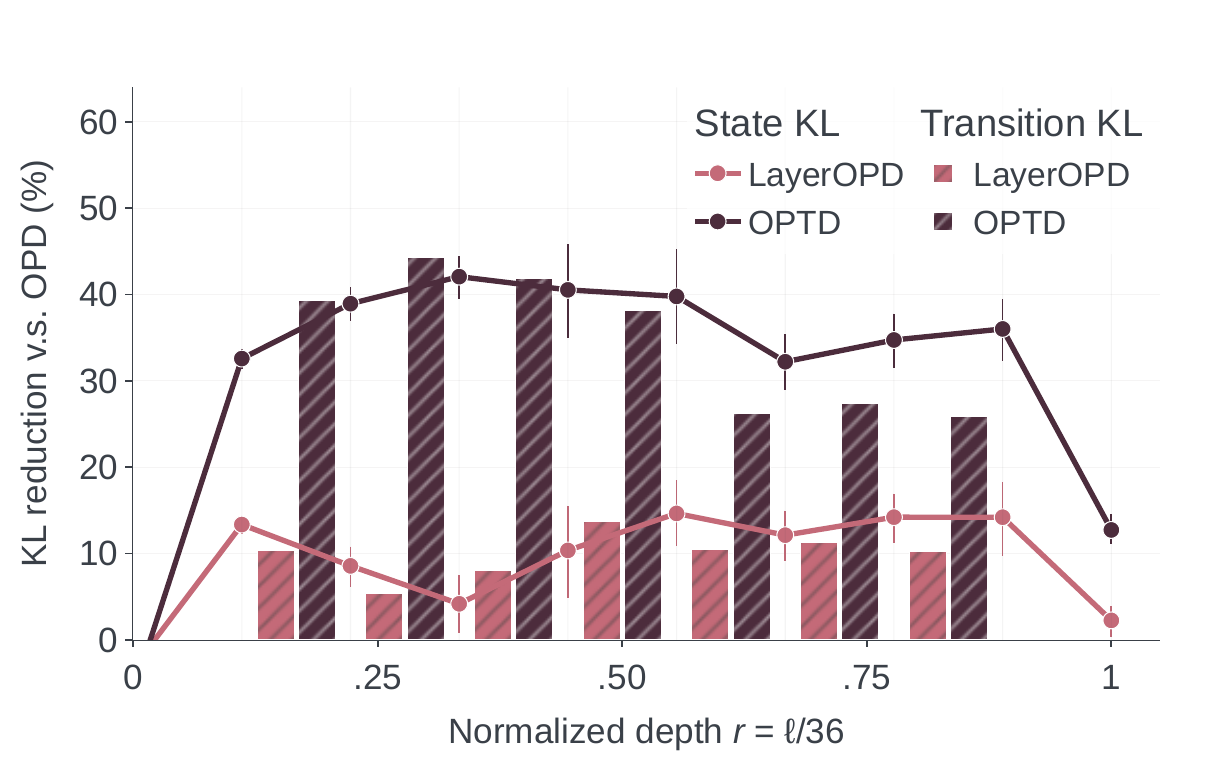}
    \caption{\textbf{Teacher alignment reductions against OPD along depth.}
        Qwen3-4B students evaluated on a shared set of saved rollout texts.
        Lines show reductions in layer-wise prediction reverse KL;
        bars show reductions in normalized-increment teacher-to-student KL between
        adjacent intermediate readouts. Horizontal axis is normalized depth
        $r=\ell/36$; error bars denote paired question-bootstrap 95\% intervals.}
    \label{fig:supervised-path-alignment}
\end{figure}

We examine whether OPTD produces the intended transition alignment and
whether this improvement also appears in layer predictions.
We compare OPD, LayerOPD, and OPTD under controlled conditions on Qwen3-4B,
evaluating all three on the same saved rollout texts.

We measure student-to-teacher KL for layer predictions (curves) and
teacher-to-student KL for normalized prediction increments (bars).
The shared-text protocol and score averaging are detailed in
App.~\ref{app:supervised-path-alignment}.
Fig.~\ref{fig:supervised-path-alignment} reports percentage reductions over OPD;
larger values indicate closer agreement with the teacher.

\begin{itemize}[leftmargin=*,noitemsep,topsep=2pt]
    \item \textbf{OPTD achieves closer transition alignment than LayerOPD.}
    Across all plotted intermediate-layer intervals, OPTD reduces transition
    KL by roughly 26\%--44\% over OPD, compared with about
    5\%--14\% for LayerOPD.
    The consistently larger reductions show that transition supervision
    brings the student's depthwise prediction revisions closer to the teacher's
    than direct alignment of layer predictions does.

    \item \textbf{Transition supervision also improves prediction alignment across depth.}
    OPTD reduces layer-wise prediction reverse KL by roughly 32\%--42\%
    at intermediate readouts and by 12.7\% at the final output.
    These reductions exceed LayerOPD's at every plotted readout,
    including its 2.2\% final-output reduction.
    Thus, supervising how predictions change across layers also improves
    agreement with the teacher's prediction states, extending the observed
    alignment benefit beyond the transitions themselves.
\end{itemize}

\noindent Together, these observations support teacher-guided prediction
revisions as an effective target for transferring knowledge across depth.

\section{Related Work}


\paragraph{On-Policy Distillation.}
Knowledge distillation transfers teacher capabilities through predictive distributions or sequences~\citep{hinton2015distilling,kim2016sequence}.
To alleviate exposure bias from offline demonstrations~\citep{bengio2015scheduled}, on-policy distillation (OPD) queries the teacher on student rollouts for dense step-level supervision~\citep{gu2024minillm,agarwal2024gkd,yang2026gopd,li2026rethinking,ma2026mopd}.
While standard OPD optimizes final-layer distributions, complementary approaches align heterogeneous intermediate hidden states through projection heads or staged schedules~\citep{romero2015fitnets,wang2020minilm,yang2026oprd,yang2026lastopd}.
However, cross-architecture matching requires careful projection designs, while static feature alignment captures proximity rather than the formation and revision of predictive beliefs across layers.

\paragraph{Depthwise Computation and Prediction Dynamics.}
A language model's final output represents the endpoint of hierarchical computations, while vocabulary lenses reveal a depthwise trajectory of hypothesis formation and refinement~\citep{belrose2023tunedlens,ali2025entropylens}.
Transformer layers introduce geometric displacements that eliminate surface-level lexical fallacies and resolve reasoning dependencies across depth~\citep{damirchi2026truthtrajectory,haim2026emergentcausal,gupta2025llmdepth}.
Recent analyses show that layer-to-layer transitions convey correctness signals separating valid reasoning from spurious correlations~\citep{damirchi2026truthtrajectory}.
However, depthwise transitions remain predominantly studied through post-hoc interpretability;
their potential as on-policy supervision signals for transferring teacher reasoning dynamics remains underexplored.

\section{Conclusion and Future Work}
\label{sec:conclusion}
We studied depthwise prediction transitions as a source of supervision for on-policy distillation.
Our diagnostic analysis links stronger teacher--student agreement in intermediate transitions to higher answer correctness, with correct responses exhibiting lower mean transition KL across all measured intermediate layer pairs.
Motivated by these observations, we proposed On-Policy Transition Distillation (OPTD), which complements final-layer distribution matching by aligning teacher--student prediction changes across depth on student-generated rollouts.
Across three backbones, three domains, and eight benchmarks, OPTD achieves the optimal overall performance in both dual- and three-domain distillation,
which demonstrates the value of depthwise transition supervision for consolidating specialist capabilities.
Future work could explore how depthwise transitions can provide supervision for general reinforcement learning.


\bibliographystyle{unsrtnat} 
\bibliography{references}


\appendix
\section{Detailed Evaluation Results}
\label{app:detailed-evaluation-results}

\subsection{Student and Teacher Reference Performance}
\label{app:reference-performance}
Tab.~\ref{tab:reference-performance} reports the initial students and their domain teachers before distillation.
Model identities are given in Tab.~\ref{tab:experiment-models}.
The mathematics scores provide the reference anchors for teacher-gap recovery.
Each specialist is reported on its corresponding domain.

\begin{table*}[t]
    \centering
    \small
    \setlength{\tabcolsep}{4.5pt}
    \begin{tabular}{@{}llrrrrrrr@{}}
        \toprule
        \multicolumn{9}{l}{\textbf{Mathematics}} \\
        \midrule
        & & \multicolumn{2}{c}{AIME24} & \multicolumn{2}{c}{AIME25}
        & \multicolumn{2}{c}{AIMO} & \\
        \cmidrule(lr){3-4}\cmidrule(lr){5-6}\cmidrule(lr){7-8}
        Size & Role & Avg & Pass & Avg & Pass
        & Avg & Pass & \\
        \midrule
        1.5B & Initial student & 30.63 & 70.00 & 21.04 & 33.33 & 63.33 & 89.16 & \\
         & Math teacher & 48.96 & 76.67 & 34.79 & 56.67 & 80.05 & 93.98 & \\
        \midrule
        4B & Initial student & 22.50 & 53.33 & 19.38 & 46.67 & 60.92 & 84.34 & \\
         & Math teacher & 53.54 & 80.00 & 41.04 & 63.33 & 79.82 & 95.18 & \\
        \midrule
        7B & Initial student & 54.58 & 86.67 & 39.38 & 70.00 & 83.13 & 96.39 & \\
         & Math teacher & 68.54 & 86.67 & 55.42 & 83.33 & 88.93 & 95.18 & \\
        \midrule
        \multicolumn{9}{l}{\textbf{Coding}} \\
        \midrule
        & & \multicolumn{2}{c}{LiveCodeBench}
        & \multicolumn{2}{c}{CodeContests} & \multicolumn{3}{c}{CodeElo} \\
        \cmidrule(lr){3-4}\cmidrule(lr){5-6}\cmidrule(lr){7-9}
        Size & Role & Avg & Pass & Avg & Pass
        & Avg & Pass & Elo \\
        \midrule
        1.5B & Initial student & 14.43 & 27.96 & 5.15 & 21.82 & 5.45 & 14.71 & 618.23 \\
        & Coding teacher & 21.86 & 36.56 & 10.11 & 23.03 & 9.38 & 22.06 & 943 \\
        \midrule
        4B & Initial student & 20.70 & 30.82 & 7.69 & 23.64 & 9.80 & 23.28 & 1112.98 \\
        & Coding teacher & 29.12 & 40.86 & 14.77 & 35.76 & 12.87 & 25.74 & 1088.54 \\
        \midrule
        7B & Initial student & 37.41 & 53.41 & -- & -- & -- & -- & -- \\
        & Coding teacher & 47.58 & 63.44 & -- & -- & -- & -- & -- \\
        \bottomrule
    \end{tabular}
    \caption{Reference performance of initial students and domain teachers.
    Accuracy scores are percentages; CodeElo is an estimated rating.
    Dashes denote unavailable reference scores.}
    \label{tab:reference-performance}
\end{table*}

\begin{table*}[t]
    \centering
    \small
    \setlength{\tabcolsep}{5pt}
    \begin{tabular}{@{}lrrrrrrrr@{}}
        \toprule
        & \multicolumn{4}{c}{IFEval} & \multicolumn{4}{c}{IFBench} \\
        \cmidrule(lr){2-5}\cmidrule(lr){6-9}
         & \makecell{strict\\prompt} & \makecell{strict\\instr.}
        & \makecell{loose\\prompt} & \makecell{loose\\instr.}
        & \makecell{strict\\prompt} & \makecell{strict\\instr.}
        & \makecell{loose\\prompt} & \makecell{loose\\instr.} \\
        \midrule
        Initial student & 39.91 & 51.72 & 43.63 & 55.46 & 12.67 & 13.66 & 15.00 & 17.73 \\
        IF teacher & 67.51 & 76.16 & 71.82 & 79.64 & 14.67 & 15.70 & 17.33 & 19.77 \\
        \bottomrule
    \end{tabular}
    \caption{Instruction-following reference accuracy (\%) for the 1.5B initial student and its domain teacher.
    Prompt accuracy requires all constraints in a prompt to pass;
    instruction accuracy scores individual constraints under strict or loose checking.}
    \label{tab:reference-ifeval}
\end{table*}

\subsection{Full Multi-Domain Results}
\label{app:multi-teacher-pipeline}
\label{app:multi-teacher-full-results}

Tabs.~\ref{tab:multidomain-math}--\ref{tab:multidomain-instruction} present the complete evaluation suite across mathematics, coding, and instruction following in the multi-domain setting.
They include detailed breakdowns for AIMO, CodeContests, CodeElo, and individual prompt/instruction criteria on IFEval and IFBench.

\begin{table*}[t]
    \centering
    \small
    \setlength{\tabcolsep}{5pt}
    \begin{tabular}{lrrrrrr}
    \toprule
    Method & \multicolumn{2}{c}{AIME24} & \multicolumn{2}{c}{AIME25} & \multicolumn{2}{c}{AIMO} \\
    \cmidrule(lr){2-3}\cmidrule(lr){4-5}\cmidrule(lr){6-7}
     & Avg & Pass & Avg & Pass & Avg & Pass \\
    \midrule
    Student & 30.63 & 70.00 & 21.04 & 33.33 & 63.33 & 89.16 \\
    Teachers & 48.96 & 76.67 & 34.79 & 56.67 & 80.05 & 93.98 \\
    OPD & 43.54 & 73.33 & 30.63 & 50.00 & \textbf{75.08} & \textbf{93.98} \\
    \rowcolor{TableGroup} \rc{recovery} & \rc{70.4\%} & \rc{49.9\%} & \rc{69.7\%} & \rc{71.4\%} & \rch{70.3\%} & \rch{100.0\%} \\
    \rowcolor{TableOurs}\textcolor{TableAccent}{\textbf{OPTD}} & \textbf{46.67} & \textbf{80.00} & \textbf{32.50} & \textbf{53.33} & 74.32 & 91.57 \\
    \rowcolor{TableOurs} \rc{recovery} & \rch{87.5\%} & \rch{149.9\%} & \rch{83.3\%} & \rch{85.7\%} & \rc{65.7\%} & \rc{50.0\%} \\
    \bottomrule
    \end{tabular}
    \caption{Mathematics results on the 1.5B student.
        Recovery is the fraction of the student--teacher gap closed by that column's score.}
    \label{tab:multidomain-math}
\end{table*}

\begin{table*}[t]
    \centering
    \small
    \setlength{\tabcolsep}{4pt}
    \begin{tabular}{lrrrrrrr}
    \toprule
    Method & \multicolumn{2}{c}{LiveCodeBench} & \multicolumn{2}{c}{CodeContests} & \multicolumn{3}{c}{CodeElo} \\
    \cmidrule(lr){2-3}\cmidrule(lr){4-5}\cmidrule(lr){6-8}
     & Avg & Pass & Avg & Pass & Avg & Pass & Elo \\
    \midrule
    Student & 14.43 & 27.96 & 5.15 & 21.82 & 5.45 & 14.71 & 618.23 \\
    Teachers & 21.86 & 36.56 & 10.11 & 23.03 & 9.38 & 22.06 & 943 \\
    OPD & \textbf{22.45} & 36.56 & 10.57 & \textbf{26.06} & \textbf{9.99} & \textbf{22.79} & 946.11 \\
    \rowcolor{TableGroup} \rc{recovery} & \rch{107.9\%} & \rc{100.0\%} & \rc{109.3\%} & \rch{350.0\%} & \rch{115.5\%} & \rch{109.9\%} & \rc{100.9\%} \\
    \rowcolor{TableOurs}\textcolor{TableAccent}{\textbf{OPTD}} & 22.13 & \textbf{36.92} & \textbf{10.72} & 25.45 & 9.59 & 20.83 & \textbf{958.12} \\
    \rowcolor{TableOurs} \rc{recovery} & \rc{103.6\%} & \rch{104.2\%} & \rch{112.3\%} & \rc{299.6\%} & \rc{105.3\%} & \rc{83.3\%} & \rch{104.6\%} \\
    \bottomrule
    \end{tabular}
    \caption{Coding results on the 1.5B student.}
    \label{tab:multidomain-code}
\end{table*}

\begin{table*}[t]
    \centering
    \small
    \setlength{\tabcolsep}{3.5pt}
    \begin{tabular}{lrrrrrrrr}
    \toprule
    Method & \multicolumn{4}{c}{IFEval} & \multicolumn{4}{c}{IFBench} \\
    \cmidrule(lr){2-5}\cmidrule(lr){6-9}
     & \makecell{strict\\prompt} & \makecell{strict\\instr.} & \makecell{loose\\prompt} & \makecell{loose\\instr.} & \makecell{strict\\prompt} & \makecell{strict\\instr.} & \makecell{loose\\prompt} & \makecell{loose\\instr.} \\
    \midrule
    Student & 39.91 & 51.72 & 43.63 & 55.46 & 12.67 & 13.66 & 15.00 & 17.73 \\
    Teachers & 67.51 & 76.16 & 71.82 & 79.64 & 14.67 & 15.70 & 17.33 & 19.77 \\
    OPD & 60.41 & 68.61 & 62.67 & 70.86 & 12.00 & 14.24 & 15.67 & 18.60 \\
    \rowcolor{TableGroup} \rc{recovery} & \rc{74.3\%} & \rc{69.1\%} & \rc{67.5\%} & \rc{63.7\%} & \rc{$-$33.5\%} & \rc{28.4\%} & \rc{28.8\%} & \rc{42.6\%} \\
    \rowcolor{TableOurs}\textcolor{TableAccent}{\textbf{OPTD}} & \textbf{60.50} & \textbf{71.04} & \textbf{63.07} & \textbf{73.27} & \textbf{13.67} & \textbf{14.83} & \textbf{17.67} & \textbf{20.06} \\
    \rowcolor{TableOurs} \rc{recovery} & \rch{74.5\%} & \rch{79.1\%} & \rch{69.0\%} & \rch{73.7\%} & \rch{50.0\%} & \rch{57.4\%} & \rch{114.6\%} & \rch{114.2\%} \\
    \bottomrule
    \end{tabular}
    \caption{Instruction-following results on the 1.5B student.}
    \label{tab:multidomain-instruction}
\end{table*}

\section{Method Implementation and Derivation}
\label{app:opd-transition-implementation}

We compare the candidate-based final-policy OPD update with the transition update used by OPTD.
Both use student-generated prefixes and fixed teacher predictions.
In the pseudocode below, \texttt{detach()} preserves a tensor's value while preventing gradients through that occurrence.
All softmax operations and candidate sums act on the last dimension.
Log-probabilities are computed over the full vocabulary before candidates are selected.

\subsection{Final-Policy OPD}

The following pseudocode shows the student-top-$k$ formulation.
The teacher is evaluated on the same candidate tokens as the student.
For readability, logits below include any configured temperature scaling.

\begingroup
\footnotesize
\begin{verbatim}
student_log_probs = log_softmax(student_logits)
teacher_log_probs = log_softmax(teacher_logits)

candidate_ids = student_logits.topk(top_k).indices
student_candidate_log_probs = gather(
    student_log_probs, candidate_ids
)
teacher_candidate_log_probs = gather(
    teacher_log_probs, candidate_ids
)

weights = softmax(
    student_candidate_log_probs.detach()
)
score_coeff = (
    weights
    * (student_candidate_log_probs.detach()
       - teacher_candidate_log_probs.detach())
).detach()

per_token_loss = (
    score_coeff * student_candidate_log_probs
).sum(dim=-1)
\end{verbatim}
\endgroup

\subsection{OPTD}

For each response position, we gather the intermediate and final prediction distributions for the same candidate tokens.
The tensors \texttt{student\_path} and \texttt{teacher\_path} have shape [batch, layer, token, candidate], with layers ordered from shallow to deep and ending at the final prediction.
Candidate identities are held fixed across layers so that each subtraction compares the same token.
The training update is:

\begingroup
\footnotesize
\begin{verbatim}
teacher_delta = (
    teacher_path[:, 1:] - teacher_path[:, :-1]
).detach()
student_delta = (
    student_path[:, 1:] - student_path[:, :-1]
)

weights = softmax(student_path[:, 1:].detach())
score_coeff = (
    weights
    * (student_delta.detach() - teacher_delta)
).detach()

per_transition_token_loss = (
    score_coeff * student_delta
).sum(dim=-1)
\end{verbatim}
\endgroup

Gradients pass through both student readouts that define each transition;
the teacher increments, candidate weights, and correction coefficients are detached.

\subsection{Loss Reduction}

Using the candidate set $\mathcal{K}_t$ from the main text, let $m_t$ denote
the response-position mask.
The loss for transition $i$ is
\[
\mathcal{L}_{\mathrm{trans}}^{(i)}
 =\frac{1}{\sum_t m_t}\sum_t m_t
   \sum_{k\in\mathcal{K}_t}
   w_{i,t}(k)\,
   \operatorname{sg}\!\left[
      \Delta^S_{i,t}(k)-\Delta^T_{i,t}(k)
   \right]\Delta^S_{i,t}(k),
\]
where $w_{i,t}$ is the destination-layer candidate weight in
Eq.~\eqref{eq:transition-weight} and is held fixed during differentiation.
In practice, evaluating the expectation in Eq.~\eqref{eq:transition-gradient} over candidate subset $\mathcal{K}_t$ restricts computation to high-probability tokens.
Token losses are reduced over the retained response positions, and the selected transitions receive equal weights:
\[
\mathcal{L}_{\mathrm{trans}}
 =\frac{1}{M}\sum_{i=1}^{M}\mathcal{L}_{\mathrm{trans}}^{(i)}.
\]
The transition update is combined with final-policy OPD;
the executed scaling is specified in App.~\ref{app:experiment-training}.
Each transition has its own weights and correction, so the objective preserves the contribution from every supervised layer pair.

\subsection{Relation to Reverse Transition KL}
\label{app:reverse-transition-kl}

The transition reward in Eq.~\eqref{eq:transition-reward} is the increment-level
counterpart of the OPD reward.
We derive the exact gradient of the
reverse transition KL and then relate it to the detached top-$K$ update used
in the main text.

Fix a response position $t$ and transition $i$.
Use $q^S$ and $q^T$ from Eq.~\eqref{eq:normalized-increment}, and write
$z^S_k=\Delta^S_{i,t}(k)$ and $z^T_k=\Delta^T_{i,t}(k)$ for brevity.
The reverse transition KL is
\[
\mathcal{K}_{i,t}=D_{\mathrm{KL}}(q^S\|q^T).
\]
Because $q^T$ is fixed, differentiating with respect to the student
increments gives
\[
\resizebox{\linewidth}{!}{$\displaystyle
\nabla_\theta \mathcal{K}_{i,t}
 =\sum_k q^S_k
 \left[
 (z^S_k-z^T_k)
 -\sum_j q^S_j(z^S_j-z^T_j)
 \right]
 \nabla_\theta z^S_k.
$}
\]
Defining the teacher-guided correction $R_k=z^T_k-z^S_k$ and its
$q^S$-weighted mean $\bar R=\sum_j q^S_jR_j$, this becomes
\begin{equation}
-\nabla_\theta \mathcal{K}_{i,t}
 =\sum_k q^S_k(R_k-\bar R)\nabla_\theta z^S_k.
\label{eq:kl-gradient-analytical}
\end{equation}
Thus, a centered reward weighted by the normalized student increments gives
the score-function form of reverse transition KL.

The practical OPTD update in Eq.~\eqref{eq:transition-gradient} uses
$w_{i,t}$ from Eq.~\eqref{eq:transition-weight}, normalized over
$\mathcal{K}_t$, and the raw transition reward
$R^{\mathrm{trans}}_{i,t}$ from Eq.~\eqref{eq:transition-reward}.
The exact reverse transition-KL gradient instead uses increment-softmax
weights $q^S$ and their centered correction.
This distinction motivates the controlled formulation comparison in \S\ref{subsec:objective-formulation-analysis}.

\section{Experimental Setup and Reproducibility}
\label{app:experiment-details}

\subsection{Datasets}
We use \href{https://huggingface.co/datasets/TURLEing/DAPO-Math-17k}{DAPO-Math-17k} for mathematics,
\href{https://huggingface.co/datasets/agentica-org/DeepCoder-Preview-Dataset}{DeepCoder-Preview-Dataset} for coding, and
\href{https://huggingface.co/datasets/TURLEing/Rubrics-To-Tokens}{Rubrics-To-Tokens} for instruction following.
Training prompts are sampled to give each domain equal representation.
Each student-generated response is assigned to its source-domain specialist.
The instruction-following teacher is added in the three-domain setting.

\subsection{Students and Domain Teachers}
\label{app:experiment-models}
Tab.~\ref{tab:experiment-models} lists the student and domain teachers for each model size.
The instruction-following teacher is added only in the multi-domain 1.5B setting.
The link icons open the corresponding Hugging Face repositories.

\newcommand{\hfmodel}[2]{#1\,\href{https://huggingface.co/#2}{\textcolor{black}{\textsuperscript{\scriptsize$\nearrow$}}}}
\begin{table*}[t]
    \centering
    \small
    \setlength{\tabcolsep}{8pt}
    \renewcommand{\arraystretch}{1.12}
    \begin{tabularx}{\textwidth}{@{}llX@{}}
        \toprule
        Size & Role & Model \\
        \midrule
        \textbf{1.5B} & Student & \hfmodel{DeepSeek-R1-Distill-Qwen-1.5B}{deepseek-ai/DeepSeek-R1-Distill-Qwen-1.5B} \\
        & Math teacher & \hfmodel{JustRL-DeepSeek-1.5B}{hbx/JustRL-DeepSeek-1.5B} \\
        & Coding teacher & \hfmodel{DeepCoder-1.5B-Preview}{agentica-org/DeepCoder-1.5B-Preview} \\
        & IF teacher & \hfmodel{Nemotron-Research-Reasoning-Qwen-1.5B (ProRL v2)}{nvidia/Nemotron-Research-Reasoning-Qwen-1.5B} \\
        \midrule
        \textbf{4B} & Student & \hfmodel{Qwen3-4B}{Qwen/Qwen3-4B} \\
        & Math teacher & \hfmodel{Qwen3-4B-Math-RL-Specialist}{pb09204048/Qwen3-4B-Math-RL-Specialist} \\
        & Coding teacher & \hfmodel{Qwen3-4B-Code-RL-Specialist}{pb09204048/Qwen3-4B-Code-RL-Specialist} \\
        \midrule
        \textbf{7B} & Student & \hfmodel{DeepSeek-R1-Distill-Qwen-7B}{deepseek-ai/DeepSeek-R1-Distill-Qwen-7B} \\
        & Math teacher & \hfmodel{Skywork-OR1-Math-7B}{Skywork/Skywork-OR1-Math-7B} \\
        & Coding teacher & \hfmodel{AceReason-Nemotron-7B}{nvidia/AceReason-Nemotron-7B} \\
        \bottomrule
    \end{tabularx}
    \caption{Students and domain teachers grouped by model size.
    The IF teacher is added only in the multi-domain extension.
    Link icons point to Hugging Face repositories.}
    \label{tab:experiment-models}
\end{table*}

\subsection{Training Configuration}
\label{app:experiment-training}

\paragraph{On-Policy Training.}
The 1.5B settings use 16 rollout prompts per update and eight prompts per
optimization minibatch;
the 4B and 7B settings use 128 and 64, respectively.
Each prompt contributes two student-generated responses, and each rollout
batch supplies two optimization minibatches.
Training runs for 150 updates at 1.5B, 50 at 4B, and 100 at 7B.
All settings use BF16 and a learning rate of $10^{-5}$, held constant after
warmup over the first 3\% of training.
Prompts are capped at 2,048 tokens;
responses are capped at 16,384 tokens for dual-domain training and 10,240
for three-domain training.

\paragraph{Supervised Depth Paths.}
Each OPTD path consists of three intermediate readouts followed by the final
output, yielding three supervised transitions.
The dual-domain comparisons use intermediate blocks $7/13/19$ and $11/15/19$
at 1.5B, $7/13/19$ and $14/19/24$ at 4B, and $13/17/21$ at 7B.
For the 1.5B student, the main-result labels Middle and Late denote
blocks $7/13/19$ and $11/15/19$, respectively.
The main three-domain configuration uses the latter, a compact Late path
with a four-block span.
In the 1.5B geometry analysis, Middle and Late denote centers at blocks
13 and 15, respectively.
For a block center $c_b$ and block span $s_b$, the selected blocks are
$(c_b-s_b,c_b,c_b+s_b)$; the normalized center and span are
$c=c_b/28$ and $s=s_b/28$.
The tested block spans are 4, 6, and 8, corresponding to normalized spans
of approximately $0.143$, $0.214$, and $0.286$.
Other paths are specified in the corresponding analyses.
All block indices are zero-based.
Intermediate next-token distributions are obtained by applying the model's
final normalization and shared output projection to the selected hidden
states, with temperature 1.

\paragraph{Transition Supervision.}
Final-policy OPD selects the student's top-16 candidate tokens at each
response prefix and evaluates the assigned teacher on the same candidates.
OPTD retains these candidate identities across the supervised path and
applies transition supervision to the last 2,000 response prediction
positions, or the entire response when it is shorter.
The three transition losses receive equal weight and are summed with the
final-policy OPD loss, corresponding to a weight of 3 on the mean transition
objective in Eq.~\eqref{eq:transition-gradient}.
The combined loss updates the student's intermediate representations and
shared output projection, with teacher predictions and reward weights held fixed.

\subsection{Baseline Implementations}
\label{app:experiment-baselines}
All baselines share OPTD's domain-specific teacher assignment and response
supervision span (App.~\ref{app:experiment-training}).
Where included, final-policy OPD follows the update in
App.~\ref{app:opd-transition-implementation}.

\paragraph{OPRD.}
OPRD~\citep{yang2026oprd} aligns $\ell_2$-normalized student and teacher
hidden representations using MSE with coefficient 1.
The 1.5B baseline supervises all transformer layers with this loss alone;
the 4B baseline uses the same intermediate layers as OPTD's Late path and
combines representation alignment with final-policy OPD.

\paragraph{LastOPD.}
LastOPD~\citep{yang2026lastopd} applies the same representation loss to the
final-layer hidden states.
Over training, its weight decreases linearly from 1 toward 0, while the
final-policy OPD weight increases from 0 toward 1.

\paragraph{LayerOPD.}
LayerOPD combines final-policy OPD with independent alignment of the
prediction distributions at three intermediate readouts.
It uses the same readout construction as OPTD and selects the student's
top-16 candidates separately at each readout for both models to score.
The 1.5B and 4B baselines use OPTD's Middle and Late paths, respectively;
the 7B baseline uses blocks $11/15/19$.

\subsection{Benchmark Evaluation}
\label{app:experiment-evaluation}

Sampled responses use BF16 and top-$p=0.95$.
In the result tables, Avg denotes mean per-problem accuracy and Pass denotes
the fraction of problems with at least one correct response; the number of
sampled responses used to estimate each metric is specified below.

\paragraph{Mathematics.}
AIME24 and AIME25 contain 30 problems each, and the AIMO evaluation subset
contains 83 problems.
We sample 16 responses per problem at temperature 0.7 with a 31,744-token
output cap and score final answers using a rule-based grader.
Math Avg. gives equal weight to the three benchmark Avg scores.
The reference scores use non-thinking prompts for Qwen3-4B and reasoning
prompts for the R1-Distill students.

\paragraph{Coding.}
LiveCodeBench v5 uses the 279-problem DeepCoder test set at temperature 0.6.
The main comparisons report Avg and Pass from eight responses per problem;
the output cap is 32,768 tokens.
The 1.5B reference scores combine the preserved four-response batch with an
independent four-response batch under the archived four-seed protocol;
the 7B reference scores use native reasoning templates.
CodeContests uses 165 test problems with 16 Python 3 solutions per problem,
temperature 0.7, and a 15,360-token output cap.
Solutions must pass all public, private, and generated tests, with a
10-second per-test timeout and a 1,024 MiB memory limit.
We report a CodeElo estimate~\citep{quan2025codeelo} on the DeepCoder
evaluation split~\citep{luo2025deepcoder}, comprising 408 Codeforces problems
across 57 contests, with eight responses per problem, temperature 0.6,
and a 32,768-token output cap.
Candidates are graded locally with SandboxFusion.
Following the CodeElo methodology (Sections 3.3 and 4.1), we compute contest
scores from the first accepted candidate and preceding failed attempts,
estimate a rating using human contest standings and pre-contest ratings,
and average the ratings across contests.\footnote{\url{https://github.com/QwenLM/CodeElo}.}
The result is an estimate because verdicts come from local tests;
the original CodeElo protocol submits solutions to the official Codeforces judge.

\paragraph{Instruction Following.}
IFEval uses all 541 prompts with 16 responses per prompt, temperature 0.7,
and a 31,744-token output cap.
IFBench uses all 300 test prompts with one greedy response per prompt and
the same output cap.
Both benchmarks report strict and loose prompt- and instruction-level
accuracies, as defined in Table~\ref{tab:reference-ifeval}; both models use
the shared rLLM reasoning template within each benchmark.

\paragraph{Teacher-Gap Recovery.}
Following the gap recovery rate in prior OPD work~\citep{li2026rethinking},
we measure the fraction of the initial student--teacher performance gap
closed by training:
\[
\mathrm{Recovery}
=100\times\frac{s_{\mathrm{trained}}-s_{\mathrm{initial}}}
{s_{\mathrm{teacher}}-s_{\mathrm{initial}}}\,\%.
\]
Here, $s_{\mathrm{initial}}$, $s_{\mathrm{trained}}$, and
$s_{\mathrm{teacher}}$ denote the scores of the initial student, trained
student, and corresponding domain teacher under the same evaluation protocol.
A recovery of $0\%$ denotes no change from the initial student;
$100\%$ denotes matching the teacher, and values above $100\%$ denote
surpassing it.
We apply this definition to each reported metric using the reference scores
in Tables~\ref{tab:reference-performance} and~\ref{tab:reference-ifeval}.
For aggregate scores such as Math Avg., we compute recovery from the
aggregate scores themselves rather than averaging per-benchmark recovery rates.

\subsection{Teacher Agreement Diagnostic}
\label{app:transition-diagnostic}

\paragraph{Data and Readouts.}
Fig.~\ref{fig:diagnostic-agreement} uses Qwen3-4B after 50 OPD updates and its mathematics specialist teacher in Tab.~\ref{tab:experiment-models}.
We retain 60 questions with both correct and incorrect answers among eight archived responses per question: 14 AIME24, 13 AIME25, and 33 AIMO questions, totaling 480 responses.
Divergences follow Eqs.~\eqref{eq:metric-final-kl} and~\eqref{eq:metric-trans-kl}, using full-vocabulary student-to-teacher KL averaged over response positions.
Intermediate predictions use each model's final normalization and unembedding head;
final readouts use the already normalized hidden state.
Figure layer labels are zero-based decoder block indices.

\paragraph{Agreement and Correctness (Panel a).}
Transition agreement is the negative arithmetic mean of transition KL over all 36 layer pairs among blocks $5,7,\ldots,21$;
final-layer agreement is negative final-output KL.
We rank each question's eight responses separately by each metric and divide them into four groups of two, from lowest to highest agreement.
For each group, we average correctness within questions, then across questions within each benchmark, and finally equally across the three benchmarks.
Shading denotes pointwise 95\% percentile intervals from 10,000 question-bootstrap resamples stratified by benchmark (seed 20261013), retaining all responses and group assignments of each sampled question.
Lines connect the observed group means.

\paragraph{Correctness Contrast across Depth (Panel b).}
We evaluate all 45 layer pairs among blocks $5,7,\ldots,21$ and the final readout.
For each pair, divergences are averaged separately over correct and incorrect responses within each question, then across questions within each benchmark, and finally equally across benchmarks.
The heatmap reports the relative gap in Eq.~\eqref{eq:relative-divergence-gap}, computed from these aggregated group means.

\subsection{Internal Alignment Evaluation}
\label{app:supervised-path-alignment}

The analysis in \S\ref{subsec:prediction-path-analysis} compares
Qwen3-4B OPD, LayerOPD, and OPTD at 50 updates under shared
token contexts.
We select 12 questions from each of AIME24, AIME25, and AIMO without
filtering by correctness or response length.
Each question contributes one OPD response and one OPTD response,
giving 36 questions and 72 texts.
All three students and the teacher process the same
prompt--response sequences, so alignment is measured at identical contexts.

We measure both layer predictions and prediction transitions over the full
vocabulary at all response token positions.
Intermediate predictions are obtained using each model's own final
normalization and LM head.
The layer-prediction curves in Fig.~\ref{fig:supervised-path-alignment}
report student-to-teacher KL,
$D_{\mathrm{KL}}(p^S_{\ell,t}\|p^T_{\ell,t})$,
at zero-based blocks $0,4,8,12,16,20,24,28,32$ and the final output.
The transition bars report teacher-to-student KL,
$D_{\mathrm{KL}}(q^T_{\ell\rightarrow\ell',t}\|q^S_{\ell\rightarrow\ell',t})$,
using the normalized increments in Eq.~\eqref{eq:normalized-increment}
between adjacent intermediate readouts from block 4 to block 32.
Readouts are positioned at normalized depth $r=\ell/36$, with the final
output placed at $r=1$.

For each readout or interval, we first average KL over response positions
within each text, then over the two response sources within each question.
We average questions within each benchmark and give the three benchmarks
equal weight.
The plotted reduction is
$100(1-\text{method distance}/\text{OPD distance})$,
computed from these aggregate distances;
positive values indicate closer teacher alignment than OPD.

The layer-prediction curves include 95\% percentile intervals from 20,000
paired question-bootstrap resamples.
Questions are resampled within each benchmark, keeping both response sources
and all students paired, and the aggregate reduction is recomputed in
each resample.

\section{Additional Analyses}
\label{app:additional-analyses}

\subsection{Final-Output Agreement Across Training}
\label{app:final-kl-training}

We examine whether the final-output pattern in \S\ref{subsec:preliminary-analysis} is already present early in training.
This supplementary analysis follows a Qwen3-4B OPD run trained on a fixed pool of 1,200 mathematics and 1,200 coding prompts, evaluating after 10, 20, 30, 40, and 50 updates.
At each stage, we use the first eight archived responses per mathematics question and all four responses per LiveCodeBench question, retaining questions with both correct and incorrect answers.
We compute full-vocabulary student-to-teacher reverse KL on student-generated prefixes, average over response tokens and then responses within each correctness group, and weight questions equally within each benchmark.

\begin{figure}[t]
    \centering
    \includegraphics[width=\linewidth]{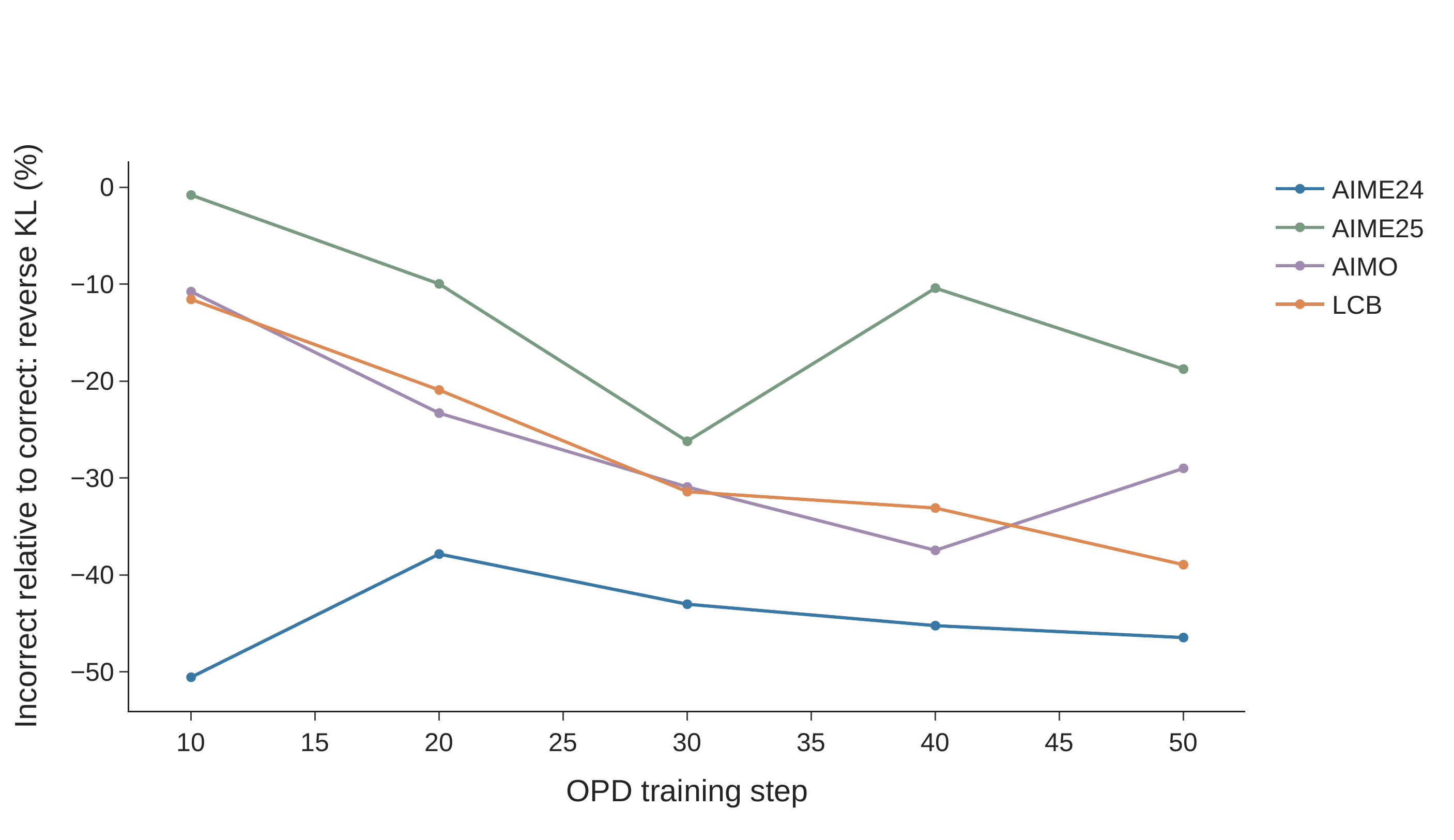}
    \caption{\textbf{Lower final-output KL for incorrect responses appears early and persists across training.}
    Each curve shows the percentage difference between the incorrect- and correct-response means, computed as $\Delta \mathcal{D}_{\mathrm{final}}$ in Eq.~\eqref{eq:relative-divergence-gap}.
    Negative values indicate closer final-output agreement for incorrect responses.
    Questions with mixed outcomes are selected separately at each training stage.}
    \label{fig:final-kl-training}
\end{figure}

Fig.~\ref{fig:final-kl-training} shows lower mean KL for incorrect responses on all four benchmarks at every evaluated stage.
The pattern is already pronounced on AIME24 after 10 updates (50.6\% lower), while AIME25 is nearly tied at this stage (0.8\% lower).
By 50 updates, incorrect responses have lower KL by 46.5\%, 18.8\%, 29.0\%, and 38.9\% on AIME24, AIME25, AIMO, and LiveCodeBench, respectively.
The size of the difference varies across benchmarks and need not increase monotonically with training.

To check whether these changes arise solely from changes in the question set, we also restrict each benchmark to questions with mixed outcomes at all five stages: 9 AIME24, 7 AIME25, 17 AIMO, and 12 LiveCodeBench questions.
On these fixed question sets, the lower-KL pattern remains present in every benchmark at every stage.
From 10 to 50 updates, the relative reduction grows from 4.4\% to 26.2\% on AIME25, from 12.4\% to 24.6\% on AIMO, and from 14.9\% to 39.4\% on LiveCodeBench; AIME24 remains pronounced at 50.7\% and 48.1\%, respectively.
These observations show that the motivating phenomenon extends across training stages, while its evolution is benchmark-dependent.

\subsection{Predictive Weights and Reward Centering}
\label{app:centered-predictive-instability}

The \textit{Predictive weights + centered reward} ablation in
\S\ref{subsec:objective-formulation-analysis} combines destination-layer
predictive weights with the transition reward centered under $q^S$.
We examine how this pairing changes the update coefficients.
For one transition and one response position, let
$R_k=z^T_k-z^S_k$ and define
\[
\bar R_q=\sum_k q^S_kR_k.
\]
The centered predictive-weighted coefficient is
\[
 C^{\mathrm{pred+center}}_k
 =-w_{i,t}(k)(R_k-\bar R_q),
\]
whereas OPTD's raw predictive-weighted coefficient is
\[
 C^{\mathrm{pred}}_k=-w_{i,t}(k)R_k.
\]
Their difference is an additional term proportional to the predictive weights:
\[
 C^{\mathrm{pred+center}}_k-C^{\mathrm{pred}}_k
 =w_{i,t}(k)\bar R_q.
\]
The reverse-KL gradient is centered under $q^S$ and therefore satisfies
$\sum_k q^S_k(R_k-\bar R_q)=0$.
The predictive-weighted coefficient instead has
\[
\resizebox{\linewidth}{!}{$\displaystyle
 \sum_kw_{i,t}(k)(R_k-\bar R_q)
 =\sum_kw_{i,t}(k)R_k-\sum_kq^S_kR_k,
$}
\]
which is generally nonzero.
Centering under $q^S$ therefore adds $w_{i,t}(k)\bar R_q$ to the
raw predictive-weighted coefficient.
The resulting coefficients do not generally retain the zero-sum property
of the reverse transition-KL gradient derived in
App.~\ref{app:reverse-transition-kl}.
This coefficient difference distinguishes the two updates;
we next examine their gradient scales during training.

An analytical and numerical backward-pass audit confirms the implemented
formulations, with maximum logit-gradient discrepancies below $10^{-16}$.
Table~\ref{tab:centered-predictive-gradient-audit} reports the recorded
gradient scales under the matched 150-update setting.

\begin{table}[t]
    \centering
    \small
    \setlength{\tabcolsep}{7pt}
    \resizebox{\columnwidth}{!}{%
    \begin{tabular}{lrr}
    \toprule
    Update branch & \makecell{Maximum\\gradient norm} & \makecell{First update\\above 100} \\
    \midrule
    \makecell[l]{Increment weights + raw reward\\($q^S$)} & 16.06 & never \\
    \makecell[l]{Predictive weights + centered reward\\($w_{i,t}$)} & 4576.0 & 34 \\
    ReverseKL ($q^S$, centered) & 3.10 & never \\
    \bottomrule
    \end{tabular}%
    }
    \caption{Gradient-scale audit over 150 updates for the three transition-update
    branches shown.
    Among these branches, only the centered predictive branch has a recorded
    gradient norm above 100, reaching 4576 before clipping.}
    \label{tab:centered-predictive-gradient-audit}
\end{table}

The centered predictive branch reaches a maximum pre-clipping gradient norm
of 4576, compared with 16.06 and 3.10 for the two increment-weighted branches.
Together with the controlled results in
Table~\ref{tab:transition-objective-ablation}, this supports pairing
predictive weights with the raw transition reward in OPTD.

\end{document}